\documentclass{article}

\usepackage{arxiv}

\usepackage[utf8]{inputenc} 
\usepackage[T1]{fontenc}    
\usepackage{hyperref}       
\usepackage{url}            
\usepackage{booktabs}       
\usepackage{amsfonts}       
\usepackage{nicefrac}       
\usepackage{microtype}      
\usepackage{lipsum}
\usepackage{graphicx}
\usepackage{amsmath}
\usepackage{float}
\usepackage{cite}
\usepackage{xcolor}

\title{Mind the gap: Challenges of deep learning approaches to Theory of Mind}
\author{
  \begin{tabular}{cc}
    \shortstack{Jaan Aru\\
   Institute of Computer Science\\
   University of Tartu\\
   \texttt{jaan.aru@gmail.com}}
 &
   \shortstack{Aqeel Labash\\
   Institute of Computer Science\\
   University of Tartu\\
  \texttt{aqeel.labash@gmail.com}} \\\\
    \shortstack{Oriol Corcoll\\
   Institute of Computer Science\\
   University of Tartu\\
  \texttt{ocorcoll@gmail.com}}
 &
   \shortstack{Raul Vicente\\
   Institute of Computer Science\\
   University of Tartu\\
  \texttt{raulvicente@gmail.com}}
\end{tabular}
}

\begin{document}
\maketitle

\begin{abstract}
Theory of Mind is an essential ability of humans to infer the mental states of others. Here we provide a coherent summary of the potential, current progress, and problems of deep learning approaches to Theory of Mind. We highlight that many current findings can be explained through shortcuts. These shortcuts arise because the tasks used to investigate Theory of Mind in deep learning systems have been too narrow. Thus, we encourage researchers to investigate Theory of Mind in complex open-ended environments. Furthermore, to inspire future deep learning systems we provide a concise overview of prior work done in humans. We further argue that when studying Theory of Mind with deep learning, the research’s main focus and contribution ought to be opening up the network’s representations. We recommend researchers to use tools from the field of interpretability of AI to study the relationship between different network components and aspects of Theory of Mind.
\end{abstract}


\section{Introduction}\label{sec1}

Rapid advances in deep learning (DL) have led to human-level performance on certain visual recognition and natural language processing tasks. Moreover, research has revealed shared computational principles in humans and DL models for vision \cite{gucclu2015deep,seibert2016performance,cichy2016comparison,yamins2016using,kriegeskorte2015deep} and language processing \cite{schrimpf2021neural, caucheteux2022brains, goldsteinshared}. In particular, there is a hierarchical correspondence between the neural activity at different levels of processing and the different layers of deep learning systems \cite{gucclu2015deep,seibert2016performance,cichy2016comparison}. This work has illuminated how relatively simple transformations applied throughout a hierarchy of processing stages can account for some characteristics of object recognition \cite{yamins2016using,kriegeskorte2015deep}. Recently, it has been demonstrated that a similar hierarchical correspondence can also be observed during natural language processing \cite{schrimpf2021neural, caucheteux2022brains, goldsteinshared}. These findings do not imply that DL has fully captured how these processes operate in the human brain. Still, DL has contributed to better characterizing the computational principles underlying them. Can DL similarly contribute to studying Theory of Mind (ToM)? 

ToM is an essential ability of humans to infer the mental states of others, such as, for example, their perceptual states, beliefs, knowledge, desires, or intentions (for review \cite{apperly2010mindreaders,heyes2014cultural, tomasello2014natural, wellman2014making}). It is a field that has been studied extensively \cite{wellman2001meta, wellman2014making, rakoczy2022foundations}, with thousands of research articles having investigated different aspects of ToM, its development \cite{apperly2010mindreaders, wellman2014making, rakoczy2022foundations} and its deficits in different psychiatric disorders such as autism \cite{apperly2010mindreaders, baron2000theory}. There are many tasks that have been designed for testing the different aspects of ToM in humans \cite{apperly2010mindreaders, quesque2020theory, rakoczy2022foundations}. For instance, one crucial milestone for ToM in human children is whether they understand that a person can have a false belief, i.e., the person believes that something is the case although in reality it is not \cite{wellman2001meta, siegal2008marvelous, wellman2014making}. However, as we will discuss below, acquiring ToM is not equivalent to passing a false-belief task, but rather is a complex skill that takes years to develop in humans \cite{tomasello2014natural, wellman2014making, wellman2020reading, rakoczy2022foundations}. For instance, a child who passes the false-belief task, is still not capable to understand that a person might pretend to believe or feel one way but actually believes or feels the opposite \cite{wellman2014making, wellman2020reading}. Given the complexity of ToM it is unclear whether any of the success of DL in vision and language tasks would carry over to ToM.

Nevertheless, it is worth investigating ToM with DL for several reasons. First, there is a pressing concern to align the strategies discovered by DL models to human needs, desires, and values \cite{leike2018scalable,christian2020alignment,kenton2021alignment}. Furthermore, a practical goal of DL systems is to build artificial agents that interact with, support, and understand humans. To achieve this, there might not be a way around studying ToM because, at least according to some prominent views about communication, ToM is necessary for the emergence of meaningful communication and language \cite{tomasello2010origins,tomasello2014natural,scott2014speaking,mercier2017enigma}. According to this perspective, the only way to build agents that can communicate with humans or each other in a meaningful fashion is to understand and develop ToM capabilities. 

Finally, given that there are still many open questions about ToM even in humans \cite{siegal2008marvelous,call2011does,apperly2010mindreaders,wellman2014making,heyes2014cultural,quesque2020theory}, modern DL tools could in principle help to understand ToM at an unprecedented level. For instance, while most researchers agree that language is important for the development of ToM \cite{siegal2008marvelous, wellman2014making}, the study of the role of language in humans has its own constraints. For example, it would be unethical to deprive children of language input. In DL models, on the other hand, one can easily parametrically change the amount of language input and precisely manipulate the words and phrases the model experiences. DL models are also useful because one can open or modify individual components of the DL architecture. For instance, in DL systems, one can find the artificial neurons and neuron populations that model other agents and then manipulate these components. This way it is possible to evaluate the necessity or sufficiency of specific elements of ToM in diverse tasks. In short, work on DL can illuminate aspects of ToM that are otherwise hard, if not impossible, to study in humans. However, this potential can be realized only if the DL models of ToM come close enough to natural ToM.

Our goal in this paper is to highlight common problems in DL-based ToM research and to provide a concise overview of prior work done in humans that could inspire future DL systems. In the next section, we start by comparing ToM to vision to demonstrate that it will be more difficult to develop systems that have ToM-like abilities. In the third section, we will consider the problem of shortcuts: current DL-systems likely develop decision-rules that do not amount to ToM-skills. In section four, we explain that ToM is not a skill that emerges from training on a specific task. Section five looks at inherent biases that humans might have to develop ToM. The sixth section is focused on how DL might learn ToM with humans in the loop. In section seven, we arrive at the question of how to evaluate ToM in future DL systems.

\section{Will Theory of Mind be more challenging than vision for DL?}\label{sec2}

Four specific aspects have enabled DL to help make advances in visual recognition and in shedding light on the processes underlying vision in the brain. 1) Appropriate training data: DL models are trained on datasets that are directly relevant for biological vision (e.g., Imagenet \cite{deng2009imagenet}). 2) Built in and learned invariances: DL systems have achieved a certain level of robustness and generalization due to the invariances implemented in the network (e.g. convolutional kernels to achieve translational invariance) or learned from the variability in which an object is presented in the training dataset; 3) DL systems have a training objective that is similar to biological vision: Machine vision has the straightforward goal to accurately recognize objects in a scene (or to segment or localize them), and it is reasonable to posit that at least one goal (if not the goal) of biological vision is also to recognize objects; 4) There exist appropriate neuroscientific comparison data: we have a reasonably good understanding of the areas involved in vision, how they are connected and organized. Decades of work on the visual system have revealed the visual processing hierarchy, which can and has been compared to the hierarchy of transformations learned by DL systems \cite{gucclu2015deep,seibert2016performance,cichy2016comparison,yamins2016using,kriegeskorte2015deep, kuzovkin2018activations}. 

In ToM research, the connection between DL and biology is much more challenging to establish. In DL, ToM is mainly studied with deep reinforcement learning (DRL), where the data that the agent experiences and the objective are intermingled. The objective is only implicitly defined via the reward structure, which drives the agent's actions. In turn, the agent's actions determine the reward that it experiences and thus create constant feedback. Hence, in DRL the reward structure of the task is a central factor that determines precisely what the agent does and learns. However, in the case of ToM, there might not exist a simple and explicit cost function or a “reward structure” that would necessarily lead to the emergence of ToM, although recent work has started to unravel the cost functions for social referential communication \cite{jara2021social}. 

The problem is further exacerbated by the fact that there is no appropriate neuroscientific comparison data. In vision, it is relatively straightforward to present the same images to humans and deep neural networks and study the correspondences between these representations \cite{gucclu2015deep,seibert2016performance,cichy2016comparison,kuzovkin2018activations}. In particular, one can compare the activation of a specific layer of the deep neural network with the activity patterns measured in the human brain and ask how much variance of the biological data is explained by the DL model. For ToM such data does not yet exist and therefore it is hard to study the correspondence. 

In addition, the datasets that are used to train machine vision algorithms have variability with regard to the exact position, angle, etc of the objects, thus enabling invariances to arise. Even more than that, DL networks that can cope with visual recognition tasks have in-built structural biases such as convolutions. In contrast, we have currently very little understanding of invariances desired (to be implemented or gained through learning) to achieve a ToM that is robust and generalizes to situations different from the trained scenarios. 

In summary, ToM is more complex than vision and thus it will be more challenging to develop systems that have ToM-like abilities. The success of DL in vision has benefited from factors that are not (yet) available for ToM.   

\section{Shortcuts in Theory of Mind tasks}\label{sec3}

\begin{table}[h]

\centering

        \begin{tabular}{|p{0.3\linewidth} | p{0.3\linewidth} | p{0.35\linewidth}|}
    \hline
    Task (Papers)&ToM-like expected solution & Potential shortcut\\
    \hline
       \hline
        Agent avoids consuming rewards observed by dominant agent \cite{labash2020perspective}&Agent learns the perspective of the dominant agent & Combination of position and orientation of dominant and subordinate in addition to food position\\
        \hline
        Agent observes a simulation of the false belief test \cite{rabinowitz2018machine,nguyen2020cognitive}& Agent learns to differentiate between internal states of other agents to predict where the observed agents will go& Agent exploits a combination of features such as positions and distances between elements \\
        \hline
        Game-theoretic tasks \cite{foerster2017learning,freire2019modeling,robert1984evolution}&Predict the other agent's policy to cooperate or defect & Tit-for-tat strategy\\
        \hline
        Three agents cooperate to cover three landmarks \cite{matiisen2018deep,lowe2017multi}&Agents learn the landmarks that other agents intend to cover&Agents exploit the initial configuration of agents and landmarks.\\
        \hline
        Hanabi card game \cite{bard2020hanabi,foerster2019bayesian,fuchs2021theory}&Agents develop beliefs about the beliefs of other agents to provide best hints & Decisions are made based on the statistical regularities \\
        \hline
        \end{tabular}
        
    \caption{\textbf{DL agents potentially learn shortcuts instead of ToM-skills.} The first column briefly specifies the task used by the researchers. The second and the third column of the table highlight the difference between the ToM-like behavior expected by the agent compared to a potential shortcut that might lead to solving the task. Note that we are not saying that in these papers the DL agents indeed used these shortcuts. We are simply pointing out that these are potential shortcuts that the agents might have used.}
    
    \label{tab1}
    
\end{table}

When one wants to develop a DL system that plays Atari or GO, one lets the system play Atari games or GO. This is not so simple in ToM: the fundamental problem is that no task unequivocally taps into ToM (as explained in the next section). Therefore, it is impossible to optimize DL systems directly for ToM. Instead, researchers use tasks that, according to their intuition, need ToM. A significant problem with this approach is that even if we humans think that one or another particular task requires ToM-like skills such as perspective taking or intention understanding, it does not mean that the DL agent indeed acquires ToM-like skills when trained on this specific task. The only pressure the agents experience is to maximise the reward on the task. If the reward can be obtained differently, without any ToM-like skills, the agent can learn to do so. 

It has been observed that DL models learn simple decision rules called shortcuts \cite{geirhos2020shortcut}. Shortcuts are tricks that enable the DL agent to get a high score on a particular task without using the processes the researchers intended the DL system to have \cite{geirhos2020shortcut, baker2020emergent, lehman2020surprising}. For instance, in object recognition, the DL system might learn to classify the objects based on a background element instead of learning the features of the target objects\cite{geirhos2020shortcut}. Given that shortcuts can arise even in vision, which is arguably simpler than ToM, then one should also expect these shortcuts in ToM, where the lack of a strong and specific pressure to infer hidden states of other agents can result in DL agents learning simpler decision rules.

Regarding ToM, this problem of a shortcut is not specific to DL. Namely, there is a long-lasting discussion about whether infants and chimpanzees possess some ToM-like skills or whether they use low-level cues (i.e., shortcuts) to solve these tasks. For instance, in experiments with chimpanzees it was observed that the subordinate animal behaved as if they could take the perspective of the dominant animal \cite{hare2000chimpanzees,hare2001chimpanzees}. In particular, in a competitive setting, the animals seemingly could infer what the other chimpanzee could see \cite{hare2000chimpanzees} and know \cite{hare2001chimpanzees}. This behavior was akin to the process of visual perspective-taking, which is an ability to appreciate the visual experience of another person and that has been thoroughly studied in human ToM research \cite{apperly2009humans,apperly2010mindreaders}. However, these results were immediately followed by discussion whether chimpanzees use low-level cues or ToM to solve tasks like this \cite{povinelli2003chimpanzee, tomasello2003chimpanzees}. Similarly, a well-known paper measured eye-movements (in the so-called anticipatory looking test) to claim that infants understand false beliefs \cite{onishi200515}. This led to an ongoing discussion about the reliability of such measures and whether they really demonstrate ToM understanding in infants \cite{ruffman2005infants, heyes2014false, rakoczy2022foundations}. When researchers used the anticipatory looking test on chimpanzees and showed that based on this measure, chimpanzees pass a false-belief task \cite{krupenye2016great}, a discussion ensued about how much of this can be traced back to ToM \cite{heyes2017apes, krupenye2017test, horschler2020non}. Even more, the same argument can be made about adult humans: under certain circumstances, for instance when resources are occupied by another task, humans use more automatic responses, general knowledge and heuristics (i.e., shortcuts) to solve ToM tasks \cite{apperly2009humans, apperly2010mindreaders}. Given that we cannot be sure whether adult humans, infants and animals with similar brains to us are using ToM or shortcuts, we should be wary of attributing ToM-like skills to artificial agents performing simpler tests. 

\begin{figure}[ht]%
\centering
\includegraphics[width=0.9\textwidth]{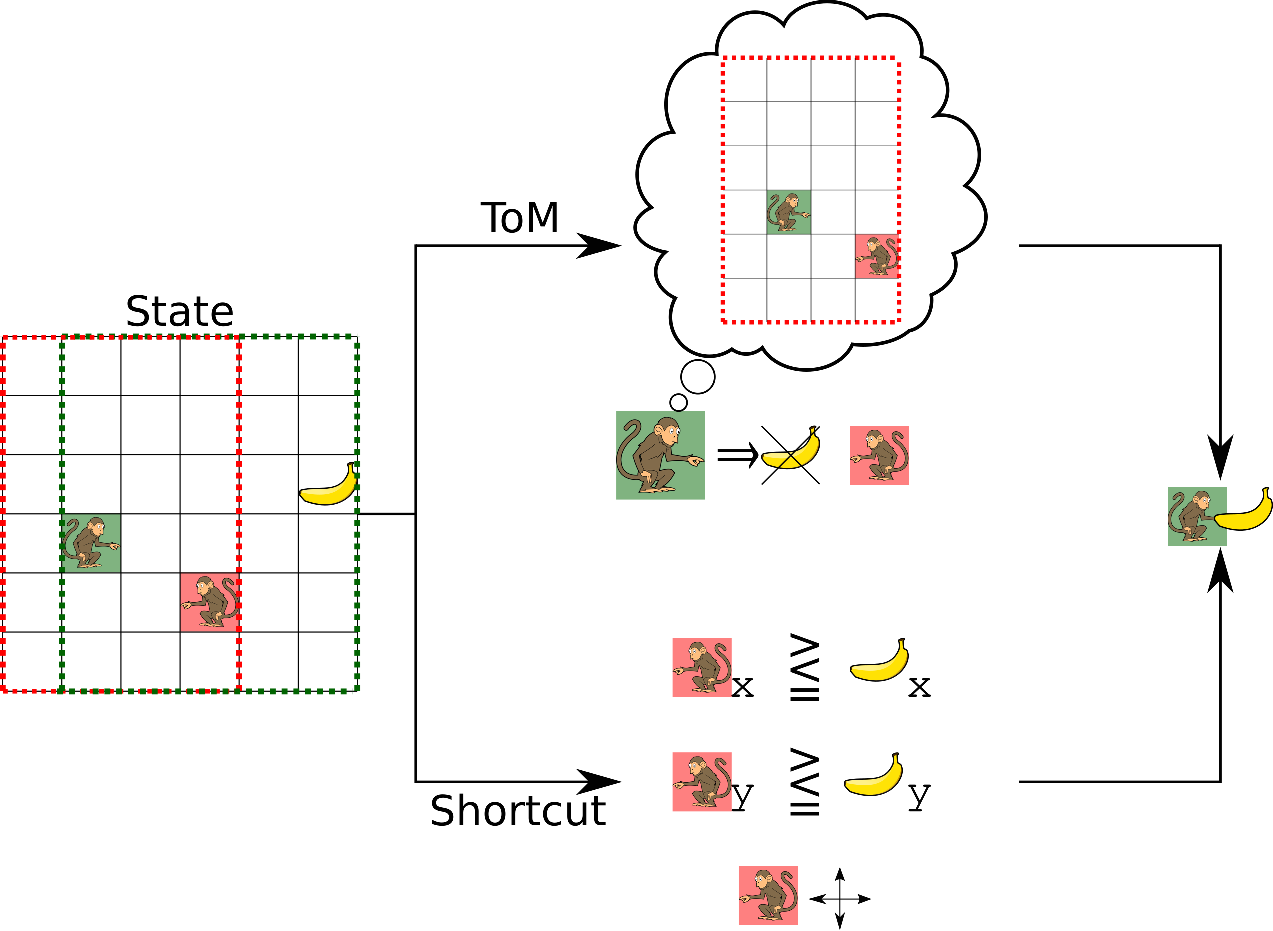}
\caption{\textbf{ToM vs shortcut in artificial agents solving a perspective-taking task.} \textbf{Left} Example of an environment state in the task. The dominant agent is marked with the red square, the subordinate agent with the green square. The visual field of the dominant agent is surrounded by red dashed lines, the visual field of the subordinate with green dashed lines. \textbf{Right, top} Based on human behavior one could think that the agent on the green square infers that the dominant agent cannot see the banana and therefore goes for the banana. \textbf{Right, bottom} However, the shortcut solution is that the agent simply takes into account the orientation and the distance of the dominant agent without inferring anything about its perspective.}\label{fig1}
\end{figure}

This means that even if the researcher intends to study ToM, the DL agents might potentially find simple shortcuts. In Table 1 we highlight several cases, where the agents potentially learned a shortcut. To bring one concrete example, \cite{labash2020perspective} were inspired by the perspective-taking experiments with chimpanzees described above \cite{hare2000chimpanzees,hare2001chimpanzees}. The authors \cite{labash2020perspective} implemented two agents (“subordinate” and “dominant”) and investigated whether the behavior of the agents revealed some rudimentary skills of perspective-taking, similar to the experiments done with chimpanzees \cite{hare2000chimpanzees,hare2001chimpanzees}. Indeed, after training a subordinate agent solved the task: go to food if the dominant is not observing; avoid the food if the dominant is observing (Fig.\ref{fig1}).
In humans, the inference of whether some item is visible from the point of view of another is usually accompanied by the possibility of imagining such a point-of-view itself \cite{apperly2010mindreaders}. Hence, in this experiment, one might assign such ToM skill to an artificial agent that successfully solves the task \cite{labash2020perspective}. Yet, a simple "shortcut" based on the position and orientation of the other agent in relation to the food position is enough to successfully determine the agent strategy (Fig.\ref{fig1}). In short, given a reward scheme and a fixed environment, there are simple geometric combinations that allow the agent to succeed without representing other agents' perceptual states. Likewise, \cite{rabinowitz2018machine,nguyen2020cognitive} used an environment and task in which the DL agent could exploit simple combinations of geometrical features such as positions and distances between elements to solve the task successfully. 

Further, in game-theoretic scenarios, it is known that there exist simple strategies such as tit-for-tat (repeat opponent’s action from the previous round) as in \cite{foerster2017learning,freire2019modeling,robert1984evolution} which can lead to significant payoffs. In games that seemingly required coordination or communication \cite{matiisen2018deep,lowe2017multi}, the initial configuration of targets and trajectories can determine the agent’s actions without the need to coordinate or infer each other goals. This problem cannot be mitigated simply by suggesting new or more complex tasks. For example, one recent prominent proposal has been that the card-game Hanabi is suitable for studying ToM with DL \cite{bard2020hanabi, foerster2019bayesian, fuchs2021theory}. However, even in Hanabi there will be statistical regularities which can be exploited by DRL agents, making it difficult to prove that ToM-like abilities indeed contributed to solving the task.

The fact that DL agents use shortcuts in ToM tasks does not imply that this work bears no relevance at all to understanding human-like ToM. This is because, as illustrated above, humans also use shortcuts in ToM tasks. In psychological ToM-research, there is an important theoretical distinction between two different systems underlying ToM-skills \cite{apperly2009humans, apperly2010mindreaders, low2012implicit, butterfill2013construct}. One is automatic, implicit and inflexible, whereas the second one, that emerges later in development, can be used in a more flexible and explicit manner \cite{apperly2009humans, apperly2010mindreaders, low2012implicit, butterfill2013construct}. 

In summary, given that the problem of shortcuts is evident even for vision, one would expect this problem to be even more prominent for ToM tasks. However, in ToM shortcuts are not only a nuisance, but rather also constitute one valid way how humans and other animals solve ToM tasks. Similarly to human ToM research, DL research into ToM could take into account that tasks can be solved via shortcuts and differentiate between different types of ToM. This would make the DL research more relevant for the study of ToM in humans and animals and would bring us closer to building algorithms that have more human-like ToM.

\section{Towards Theory of Mind: Beyond a task}\label{sec4}
All these previously highlighted works have tried to understand ToM as a skill that can be learned based on some particular task. However, perhaps ToM is not a skill emerging from some task. It is not clear whether there exists a simple and explicit cost function or set of rewards for biological systems that would necessarily lead to the emergence of ToM. It could be a complex cost function that cannot simply be optimized by training on a specific task. 

In ToM research done on humans, there are tasks like the classic "Sally-Anne" task used to measure ToM.  In this task, the child is presented with two dolls (Sally and Anne) that enact a scene wherein Sally hides a marble in the room before leaving, after which Anne removes it from its location. ToM is displayed when the child understands Sally's false belief that the marble is still where she left it (when Sally returns). This task requires understanding that people behave according to what they believe, even if that belief is not aligned with reality. Thus, other people may be holding a false belief based on their own perspective. However, nobody would claim that children learn ToM by a repeated confrontation with the "Sally-Anne" task. ToM is not inherent to a task; it is a specific way humans deal with these types of situations. So, it cannot be assumed that simply because a task might require ToM in humans, it does so also in DRL agents. They might solve it differently, by learning a shortcut. To humans, Sally-Anne-like tasks serve as an evaluation platform of a larger and more complex system trained independently in an open-ended fashion. 

Open-endedness departs from the single-task paradigm to an unbounded number of tasks (or even no task at all, simply a world with different possibilities). To the study of ToM, open-ended environments could provide a fruitful playground where agents coordinate, cooperate and compete to solve tasks and, possibly, learn similar strategies to ToM in humans. Like human children, DL agents might need to be and learn in an open-ended environment, where ToM skills are necessary and might be acquired through interaction with other agents. Recent work \cite{guss2019minerl,wang2020enhanced,samvelyan2021minihack,team2021open,hafner2021benchmarking} has shown how powerful open-endedness can be for learning complex behavior. Particularly, \cite{team2021open} introduces XLand, a vast environment where multiple agents learn from a spectrum of completely cooperative to fully competitive tasks. Agents trained on XLand learn complex strategies to solve any given task, but it is unknown whether ToM is one of these strategies. We encourage researchers to study whether the learning of ToM-like strategies can emerge from complex environments such as XLand.

Taken together, although there are tasks that are used to measure ToM in humans, there are reasons to think that they are not suitable as benchmarks for DL systems. Open-ended environments might be preferable to develop DL agents that have ToM-like capabilities.

\section{Towards Theory of Mind: Which biases are needed?}\label{sec5}
It is important to note, we are not claiming that ToM-like skills would “pop out” from DL agents playing in open-ended environments just like that. Developing ToM requires an open-ended environment, but it might take more than simply better data to acquire ToM. Specifically, there might be several inductive biases and constraints in the human brain which enable acquiring ToM. Many of these biases are still unknown, but here we list some of the possible venues for exploration.

First, there could be be biases for attention. For example, recognizing and distinguishing other human beings is important and hence there is an innate bias for attending to faces \cite{johnson2005subcortical,reid2017human}. In particular, preferences for faces over similarly configured non-face objects are present in neonatal infants \cite{farroni2005newborns} and even in fetuses in the third trimester of pregnancy \cite{reid2017human}. Similar biases likely exist for drawing an infant's attention to speech \cite{perszyk2018linking}, hands \cite{ullman2012simple}, eyes, and gaze-direction \cite{grossmann2017eyes}, and biological motion \cite{simion2008predisposition}. These early biases make sure that the child learns about the aspects of the world that are informative about the minds of other persons. 

Second, some of these biases might be structural. For example, the human brain has special circuits devoted to ToM – areas, where activity is selectively evoked by tasks that involve considering the minds of other people \cite{saxe2013theory,koster2013decoding}. It is currently unknown whether these circuits underlie some computations or structural biases that are specific to ToM. Similarly to convolutions – structural building blocks in convolutional neural networks that help to achieve translational invariance in visual recognition tasks – these constraints would assist the human brain to extract the features and the information relevant for ToM. The trouble is that we have a very limited understanding of what these structural biases might even be. Importantly, this question might be better tackled with the tools from DL than with methods from cognitive science and neuroscience.  

Third, research on human children has described certain steps along the way of developing full-blown ToM \cite{wellman2004scaling,wellman2014making,wellman2020reading}. First, a child needs to understand that other people can have diverse desires (i.e., people desire different things). Next, a child comes to understand diverse beliefs (i.e., people have different beliefs, even about the same situation). According to this well-established framework, the third stage is the state of knowledge-access (where the child understands that “something can be true, but someone without access to it would be ignorant of it”, \cite{wellman2014making}). Only the fourth step is the stage where the child understands false belief (i.e., the child knows that something is true, but is aware that someone else might believe something different). Finally, the authors describe a state called hidden emotion \cite{wellman2014making} or hidden mind \cite{wellman2020reading}, according to which the child understands that desires, beliefs, and knowledge (i.e., internal states) might not be apparent in a person’s behavior.

This does not mean that DL definitely has to emulate these steps to acquire ToM. However, it might be informative to keep in mind that developing ToM takes time and usually progresses along this sequence in humans. One possibility is that these stages might constitute a curriculum for training DRL agents\cite{matiisen2019teacher,forestier2017intrinsically}. Alternatively, these steps might simply constitute points of comparison and useful benchmarks. Also, the progression of these steps might also indicate something about how easy or complex it is to acquire that stage, not only in humans but also in DL models. In particular, it is relatively easy to learn the stage of “diverse desires”, because desires are clearly evident in behavior and thus could be learned from visual input. On the contrary, it seems very hard if not impossible to learn the last step (hidden mind states) from visual input alone, as the overt behavior is dissociated from the internal states in the cases relevant to this step. 

Based on the human ToM research one can say that children represent the minds of others and they acquire it in a step-by-step manner. In other words, at least some aspects of the representation of the minds of others are learned. Furthermore, we also know that one crucial input for acquiring full-blown ToM is language \cite{milligan2007language,hale2003influence,wellman2014making,wellman2020reading,peterson2005steps}. For instance, deaf children born to parents who do not master sign language develop ToM much later and their development is dependent on learning sign language \cite{peterson2005steps,wellman2014making,wellman2020reading}. Furthermore, a recent meta-analysis demonstrated that language training enhances performance in ToM tasks \cite{hofmann2016training}. Currently, there is no evidence that the later steps of ToM (i.e. understanding of false belief and hidden mind states) can develop without language input. Thus, DL agents in these open-ended environments might need to be combined with the capabilities of large language models \cite{devlin2018bert,radford2019language,brown2020language,fedus2021switch}. Here also lies one potential avenue for DL to significantly contribute to the ToM research. Namely, the assumption in the research on human ToM is that specific linguistic input drives the development of ToM skills. For instance, using verbs that relate to specific mental states (e.g. “know", “think", “believe") could underlie the understanding of these mental states \cite{hofmann2016training, rakoczy2022foundations}. One could provide the DL models the exact or at least a similar language input as infants and young children get and manipulate the amount of ToM-specific linguistic input. In DL models one can not only establish the importance of language input, which most ToM-researchers would agree with, but specifically investigate which type of linguistic input (e.g., which exact phrases) drive specific ToM-skills. 

Finally, there are still many unknowns about ToM in humans \cite{apperly2010mindreaders,wellman2014making, heyes2014cultural,quesque2020theory}. In particular, the neural and computational basis of ToM is relatively unknown and unexplored (but see \cite{baker2017rational,jara2019theory}). Historically, one research direction has been about explaining ToM through the activity of so-called mirror neurons (\cite{gallese1998mirror,rizzolatti2008mirrors}). However, strong criticisms of this view (\cite{hickok2014myth,heyes2010mirror}) have curbed the enthusiasm of these early claims and the relative interest in mirror neurons has decreased considerably \cite{heyes2022happened}. As explained below, we see such controversies as an opportunity for DL researchers.

In summary, ToM is unlikely to simply arise from large amounts of training data. Rather, human brains likely have specific inductive biases that enable them to acquire ToM. The goal of future research is to understand these biases and implement them in DL systems.

\section{Theory of Mind with humans in the loop}\label{sec6}
Instead of waiting for these biases to emerge from open interactions with other agents or for researchers to establish the correct structural biases needed for ToM, it might also be possible for humans to direct the DRL agents to learn the required biases for ToM. One aspiration in DRL is to learn from experts; these can be humans or even other DRL agents. Traditionally, methods like imitation learning (IL) \cite{Pomerleau1991, Schaal1999} or inverse reinforcement learning (IRL) \cite{Ng00algorithmsfor} have been used to learn a policy or reward function, respectively. These methods require experts to demonstrate the desired behavior, which can be incredibly hard, arduous, or impossible in many interesting cases. Recent research \cite{christiano2017deep, wu2021recursively, lampinen2021tell, abramson2021imitating, deepmindinteractiveagentsteam2022creating} has proposed variations of these methods capable of producing complex agents.

A technique that has been shown to scale with the complexity of the desired behavior is learning from human preferences \cite{christiano2017deep}. This method does not rely on demonstrations by human experts; instead, this approach requires humans to rate some of the agent's trajectories and trains a reward function via supervised learning using these ratings. In contrast to more traditional reward schemes based on handcrafted rules, the learned reward function evolves with the agent's learned behavior in a curriculum-learning fashion and, to some extent, reduces the description of the desired behavior to a simple rating.

A similar but arguably more scalable approach to learning from humans is explanation-based learning. For instance, \cite{lampinen2021tell} propose to use explanations as an auxiliary task. In other words, the agent needs to explain its behavior using natural language. Although the authors use synthetic explanations, their study shows that explanations can bias the agent towards using features with more generalization power and, at least in some cases, identify the world's causal structure. 

If ToM is not an emergent property of DRL agents, maybe aspects of ToM can be incorporated through explanations. Recent work \cite{akula2022cx} takes a step in this direction by proposing an image classification system that iteratively explains its decisions to a human in hope of finding better decisions. Research has also demonstrated that real-time visualizations of the robot’s internal decisions were the most effective means for promoting human trust in robot's behavior \cite{edmonds2019tale}.

Taken together, these approaches could provide a natural platform to facilitate the learning of human-like ToM by the DL agents. Human ratings and explanations could encourage finding solutions that require ToM.

\section{Evaluating Theory of Mind in DL agents}\label{sec7}
As of now, ToM in DL is usually evaluated through the performance of some task. However, as we have highlighted in this paper, DL systems make use of shortcuts \cite{geirhos2020shortcut}, i.e., they learn to utilize decision rules that are simpler than the ones intended by the researchers. 
 
We propose that when studying ToM with DL, the primary focus and main contribution to ToM research could be opening up the network’s representations. For instance, research on ToM through DL could provide insights on the debated role of mirror neurons \cite{heyes2010mirror,hickok2014myth}. Does something akin to mirror neurons arise in the DL systems trained in open-ended environments? What happens if one would modify or remove these neurons from the system. Tackling these questions would be informative to all researchers studying ToM. Yet, up to now, most of the work done in DL that examines ToM has not inspected the trained model weights, activations, or correlation with specific aspects of ToM (see \cite{triesch2007emergence} for an early exception). If ToM representations of hidden variables of other agents are developed and used by the network, it should be possible to isolate such representations. 

Recently, the field of interpretability has received much attention from the research community \cite{du2019techniques,olah2018building,molnar2020interpretable,alharbi2021learning}, leading to novel and powerful methods that enable a better understanding of the learned representations. We recommend taking advantage of these developments for the study of ToM.

In particular, we bring to the reader’s attention the work done by \cite{hilton2020understanding} where the authors proposed to combine multiple interpretability techniques like feature visualization, attribution, and dimensionality reduction to understand vision in DRL agents. By applying these techniques to the agent’s neural network, they could identify failure cases like hallucinations of reward-leading features or even make the agent “blind” to specific high-level features revealing how relevant these are to the task.

More straightforward methods like linear probing \cite{alain2016understanding} can also be used to test if the explicit representations on the intermediate layers of the network encode some aspects of ToM. After a specific network component has been located that seems responsible for some aspect of ToM, we recommend performing ablation \cite{meyes2019ablation} of these parts of the network to check the necessity of those components to the task. If this network component is removed from the agent, will it still solve the task? If it can still perform the task, it is likely that the DL agent was using shortcuts or that this particular component is not required for performing the task.

Applying similar methodologies to agents trained in cooperative and/or competitive environments like XLand \cite{team2021open}, Hide and Seek \cite{baker2020emergent} or Capture the Flag \cite{jaderberg2019human} could bring us closer to understanding ToM-like capabilities. 

In summary, in DL systems it is possible to study the mechanisms of ToM in an unprecedented manner. However, to realize this potential, the environments, tasks and inductive biases in the DL systems need to support the emergence of ToM in DL agents.

\section{Conclusion}\label{sec8}
Theory of Mind is a central facet of human intelligence \cite{apperly2010mindreaders,tomasello2010origins,tomasello2014natural,heyes2014cultural,scott2014speaking,mercier2017enigma}. Inspired by the success of DL in understanding biological vision \cite{gucclu2015deep,seibert2016performance,cichy2016comparison,kriegeskorte2015deep,yamins2016using} and language processing \cite{schrimpf2021neural, caucheteux2022brains, goldsteinshared}, over the last years a challenge has emerged to develop DL agents that can mimic aspects of ToM. 

In this paper, we surveyed papers that have investigated ToM with DL. We have observed that DL models can develop shortcuts which means that although the researcher intends the DL system to learn ToM, the system actually might learn a much simpler decision rule. This is a problem and a challenge, but also an opportunity for future research into ToM.

DL architectures and their learning algorithms are not the ultimate brain-like learning system, but they do provide scientific models \cite{cichy2019deep} that can guide our understanding of higher mental functions such as ToM. So far, DL remains our best source to understand the working algorithms in tasks similar to those that animals have to solve. Given the difficulty of monitoring all the relevant variables in real brains, the fact that we can open these artificial algorithms and analyze them in detail provides a source of inspiration that we can not afford to leave unexplored.

\section*{Declarations}
\subsection{Funding}
This research was supported by the European Regional Funds through the IT Academy Programme, the Estonian Research Council grants PSG728 and PRG1604, the European Union’s Horizon 2020 research and innovation programme under grant agreement No 952060 (Trust AI), the Estonian Centre of Excellence in IT (EXCITE) project number TK148, and the project CardioStressCI (ERA-CVD-JTC2020-015) from the European Union's ERA-CVD Joint Transnational Call 2020. 

\subsection{Conflict of interest/Competing interests}
We declare that none of the authors have competing financial or non-financial interests as defined by Nature Portfolio.
\subsection{Acknowledgements}
We are thankful to Daniel Majoral, Roman Ring, Kadi Tulver and four anonymous reviewers for constructive comments on the manuscript.

\bibliographystyle{unsrt}  
\bibliography{sn-bibliography}  

\begin{thebibliography}{100}

\bibitem{gucclu2015deep}
Umut G{\"u}{\c{c}}l{\"u} and Marcel~AJ van Gerven.
\newblock Deep neural networks reveal a gradient in the complexity of neural
  representations across the ventral stream.
\newblock {\em Journal of Neuroscience}, 35(27):10005--10014, 2015.

\bibitem{seibert2016performance}
Darren Seibert, Daniel Yamins, Diego Ardila, Ha~Hong, James~J DiCarlo, and
  Justin~L Gardner.
\newblock A performance-optimized model of neural responses across the ventral
  visual stream.
\newblock {\em bioRxiv}, page 036475, 2016.

\bibitem{cichy2016comparison}
Radoslaw~Martin Cichy, Aditya Khosla, Dimitrios Pantazis, Antonio Torralba, and
  Aude Oliva.
\newblock Comparison of deep neural networks to spatio-temporal cortical
  dynamics of human visual object recognition reveals hierarchical
  correspondence.
\newblock {\em Scientific reports}, 6(1):1--13, 2016.

\bibitem{yamins2016using}
Daniel~LK Yamins and James~J DiCarlo.
\newblock Using goal-driven deep learning models to understand sensory cortex.
\newblock {\em Nature neuroscience}, 19(3):356--365, 2016.

\bibitem{kriegeskorte2015deep}
Nikolaus Kriegeskorte.
\newblock Deep neural networks: a new framework for modeling biological vision
  and brain information processing.
\newblock {\em Annual review of vision science}, 1:417--446, 2015.

\bibitem{schrimpf2021neural}
Martin Schrimpf, Idan~Asher Blank, Greta Tuckute, Carina Kauf, Eghbal~A
  Hosseini, Nancy Kanwisher, Joshua~B Tenenbaum, and Evelina Fedorenko.
\newblock The neural architecture of language: Integrative modeling converges
  on predictive processing.
\newblock {\em Proceedings of the National Academy of Sciences}, 118(45), 2021.

\bibitem{caucheteux2022brains}
Charlotte Caucheteux and Jean-R{\'e}mi King.
\newblock Brains and algorithms partially converge in natural language
  processing.
\newblock {\em Communications Biology}, 5(1):1--10, 2022.

\bibitem{goldsteinshared}
Ariel Goldstein, Zaid Zada, Eliav Buchnik, Mariano Schain, Amy Price, Bobbi
  Aubrey, Samuel~A Nastase, Amir Feder, Dotan Emanuel, Alon Cohen, et~al.
\newblock Shared computational principles for language processing in humans and
  deep language models.
\newblock {\em Nature Neuroscience}, 2022.

\bibitem{apperly2010mindreaders}
Ian Apperly.
\newblock {\em Mindreaders: the cognitive basis of" theory of mind"}.
\newblock Psychology Press, 2010.

\bibitem{heyes2014cultural}
Cecilia~M Heyes and Chris~D Frith.
\newblock The cultural evolution of mind reading.
\newblock {\em Science}, 344(6190), 2014.

\bibitem{tomasello2014natural}
Michael Tomasello.
\newblock {\em A natural history of human thinking}.
\newblock Harvard University Press, 2014.

\bibitem{wellman2014making}
Henry~M Wellman.
\newblock {\em Making minds: How theory of mind develops}.
\newblock Oxford University Press, 2014.

\bibitem{wellman2001meta}
Henry~M Wellman, David Cross, and Julanne Watson.
\newblock Meta-analysis of theory-of-mind development: The truth about false
  belief.
\newblock {\em Child Development}, 72(3):655--684, 2001.

\bibitem{rakoczy2022foundations}
Hannes Rakoczy.
\newblock Foundations of theory of mind and its development in early childhood.
\newblock {\em Nature Reviews Psychology}, 1(4):223--235, 2022.

\bibitem{baron2000theory}
Simon Baron-Cohen.
\newblock Theory of mind and autism: A review.
\newblock {\em International Review of Research in Mental Retardation},
  23:169--184, 2000.

\bibitem{quesque2020theory}
Fran{\c{c}}ois Quesque and Yves Rossetti.
\newblock What do theory-of-mind tasks actually measure? theory and practice.
\newblock {\em Perspectives on Psychological Science}, 15(2):384--396, 2020.

\bibitem{siegal2008marvelous}
Michael Siegal.
\newblock {\em Marvelous minds: The discovery of what children know}.
\newblock Oxford University Press, USA, 2008.

\bibitem{wellman2020reading}
Henry Wellman.
\newblock {\em Reading minds: How childhood teaches us to understand people}.
\newblock Oxford University Press, USA, 2020.

\bibitem{leike2018scalable}
Jan Leike, David Krueger, Tom Everitt, Miljan Martic, Vishal Maini, and Shane
  Legg.
\newblock Scalable agent alignment via reward modeling: a research direction.
\newblock {\em arXiv preprint arXiv:1811.07871}, 2018.

\bibitem{christian2020alignment}
Brian Christian.
\newblock {\em The alignment problem: Machine learning and human values}.
\newblock WW Norton \& Company, 2020.

\bibitem{kenton2021alignment}
Zachary Kenton, Tom Everitt, Laura Weidinger, Iason Gabriel, Vladimir Mikulik,
  and Geoffrey Irving.
\newblock Alignment of language agents.
\newblock {\em arXiv preprint arXiv:2103.14659}, 2021.

\bibitem{tomasello2010origins}
Michael Tomasello.
\newblock {\em Origins of human communication}.
\newblock MIT press, 2010.

\bibitem{scott2014speaking}
Thom Scott-Phillips.
\newblock {\em Speaking our minds: Why human communication is different, and
  how language evolved to make it special}.
\newblock Macmillan International Higher Education, 2014.

\bibitem{mercier2017enigma}
Hugo Mercier and Dan Sperber.
\newblock {\em The enigma of reason}.
\newblock Harvard University Press, 2017.

\bibitem{call2011does}
Josep Call and Michael Tomasello.
\newblock Does the chimpanzee have a theory of mind? 30 years later.
\newblock {\em Human Nature and Self Design}, pages 83--96, 2011.

\bibitem{deng2009imagenet}
Jia Deng, Wei Dong, Richard Socher, Li-Jia Li, Kai Li, and Li~Fei-Fei.
\newblock Imagenet: A large-scale hierarchical image database.
\newblock In {\em 2009 IEEE conference on computer vision and pattern
  recognition}, pages 248--255. IEEE, 2009.

\bibitem{kuzovkin2018activations}
Ilya Kuzovkin, Raul Vicente, Mathilde Petton, Jean-Philippe Lachaux, Monica
  Baciu, Philippe Kahane, Sylvain Rheims, Juan~R Vidal, and Jaan Aru.
\newblock Activations of deep convolutional neural networks are aligned with
  gamma band activity of human visual cortex.
\newblock {\em Communications biology}, 1(1):1--12, 2018.

\bibitem{jara2021social}
Julian Jara-Ettinger and Paula Rubio-Fernandez.
\newblock The social basis of referential communication: Speakers construct
  physical reference based on listeners’ expected visual search.
\newblock {\em Psychological Review}, 2021.

\bibitem{labash2020perspective}
Aqeel Labash, Jaan Aru, Tambet Matiisen, Ardi Tampuu, and Raul Vicente.
\newblock Perspective taking in deep reinforcement learning agents.
\newblock {\em Frontiers in Computational Neuroscience}, 14:69, 2020.

\bibitem{rabinowitz2018machine}
Neil Rabinowitz, Frank Perbet, Francis Song, Chiyuan Zhang, SM~Ali Eslami, and
  Matthew Botvinick.
\newblock Machine theory of mind.
\newblock In {\em International conference on machine learning}, pages
  4218--4227. PMLR, 2018.

\bibitem{nguyen2020cognitive}
Thuy~Ngoc Nguyen and Cleotilde Gonzalez.
\newblock Cognitive machine theory of mind.
\newblock In {\em Proceedings of the 42nd annual meeting of the cognitive
  science society (cogsci 2020)}, 2020.

\bibitem{foerster2017learning}
Jakob~N Foerster, Richard~Y Chen, Maruan Al-Shedivat, Shimon Whiteson, Pieter
  Abbeel, and Igor Mordatch.
\newblock Learning with opponent-learning awareness.
\newblock {\em arXiv preprint arXiv:1709.04326}, 2017.

\bibitem{freire2019modeling}
Ismael~T Freire, Xerxes~D Arsiwalla, Jordi-Ysard Puigb{\`o}, and Paul
  Verschure.
\newblock Modeling theory of mind in multi-agent games using adaptive feedback
  control.
\newblock {\em arXiv preprint arXiv:1905.13225}, 2019.

\bibitem{robert1984evolution}
Axelrod Robert et~al.
\newblock The evolution of cooperation, 1984.

\bibitem{matiisen2018deep}
Tambet Matiisen, Aqeel Labash, Daniel Majoral, Jaan Aru, and Raul Vicente.
\newblock Do deep reinforcement learning agents model intentions?
\newblock {\em arXiv preprint arXiv:1805.06020}, 2018.

\bibitem{lowe2017multi}
Ryan Lowe, Yi~I Wu, Aviv Tamar, Jean Harb, OpenAI Pieter~Abbeel, and Igor
  Mordatch.
\newblock Multi-agent actor-critic for mixed cooperative-competitive
  environments.
\newblock {\em Advances in Neural Information Processing Systems}, 30, 2017.

\bibitem{bard2020hanabi}
Nolan Bard, Jakob~N Foerster, Sarath Chandar, Neil Burch, Marc Lanctot,
  H~Francis Song, Emilio Parisotto, Vincent Dumoulin, Subhodeep Moitra, Edward
  Hughes, et~al.
\newblock The hanabi challenge: A new frontier for ai research.
\newblock {\em Artificial Intelligence}, 280:103216, 2020.

\bibitem{foerster2019bayesian}
Jakob Foerster, Francis Song, Edward Hughes, Neil Burch, Iain Dunning, Shimon
  Whiteson, Matthew Botvinick, and Michael Bowling.
\newblock Bayesian action decoder for deep multi-agent reinforcement learning.
\newblock In {\em International Conference on Machine Learning}, pages
  1942--1951. PMLR, 2019.

\bibitem{fuchs2021theory}
Andrew Fuchs, Michael Walton, Theresa Chadwick, and Doug Lange.
\newblock Theory of mind for deep reinforcement learning in hanabi.
\newblock {\em arXiv preprint arXiv:2101.09328}, 2021.

\bibitem{geirhos2020shortcut}
Robert Geirhos, J{\"o}rn-Henrik Jacobsen, Claudio Michaelis, Richard Zemel,
  Wieland Brendel, Matthias Bethge, and Felix~A Wichmann.
\newblock Shortcut learning in deep neural networks.
\newblock {\em Nature Machine Intelligence}, 2(11):665--673, 2020.

\bibitem{baker2020emergent}
Bowen Baker.
\newblock Emergent reciprocity and team formation from randomized uncertain
  social preferences.
\newblock {\em Advances in Neural Information Processing Systems},
  33:15786--15799, 2020.

\bibitem{lehman2020surprising}
Joel Lehman, Jeff Clune, Dusan Misevic, Christoph Adami, Lee Altenberg, Julie
  Beaulieu, Peter~J Bentley, Samuel Bernard, Guillaume Beslon, David~M Bryson,
  et~al.
\newblock The surprising creativity of digital evolution: A collection of
  anecdotes from the evolutionary computation and artificial life research
  communities.
\newblock {\em Artificial Life}, 26(2):274--306, 2020.

\bibitem{hare2000chimpanzees}
Brian Hare, Josep Call, Bryan Agnetta, and Michael Tomasello.
\newblock Chimpanzees know what conspecifics do and do not see.
\newblock {\em Animal Behaviour}, 59(4):771--785, 2000.

\bibitem{hare2001chimpanzees}
Brian Hare, Josep Call, and Michael Tomasello.
\newblock Do chimpanzees know what conspecifics know?
\newblock {\em Animal behaviour}, 61(1):139--151, 2001.

\bibitem{apperly2009humans}
Ian~A Apperly and Stephen~A Butterfill.
\newblock Do humans have two systems to track beliefs and belief-like states?
\newblock {\em Psychological Review}, 116(4):953, 2009.

\bibitem{povinelli2003chimpanzee}
Daniel~J Povinelli and Jennifer Vonk.
\newblock Chimpanzee minds: suspiciously human?
\newblock {\em Trends in Cognitive Sciences}, 7(4):157--160, 2003.

\bibitem{tomasello2003chimpanzees}
Michael Tomasello, Josep Call, and Brian Hare.
\newblock Chimpanzees understand psychological states--the question is which
  ones and to what extent.
\newblock {\em Trends in cognitive sciences}, 7(4):153--156, 2003.

\bibitem{onishi200515}
Kristine~H Onishi and Ren{\'e}e Baillargeon.
\newblock Do 15-month-old infants understand false beliefs?
\newblock {\em Science}, 308(5719):255--258, 2005.

\bibitem{ruffman2005infants}
Ted Ruffman and Josef Perner.
\newblock Do infants really understand false belief?
\newblock {\em Cogn. Dev}, 9:377--395, 2005.

\bibitem{heyes2014false}
Cecilia Heyes.
\newblock False belief in infancy: A fresh look.
\newblock {\em Developmental Science}, 17(5):647--659, 2014.

\bibitem{krupenye2016great}
Christopher Krupenye, Fumihiro Kano, Satoshi Hirata, Josep Call, and Michael
  Tomasello.
\newblock Great apes anticipate that other individuals will act according to
  false beliefs.
\newblock {\em Science}, 354(6308):110--114, 2016.

\bibitem{heyes2017apes}
Cecilia Heyes.
\newblock Apes submentalise.
\newblock {\em Trends in Cognitive Sciences}, 21(1):1--2, 2017.

\bibitem{krupenye2017test}
Christopher Krupenye, Fumihiro Kano, Satoshi Hirata, Josep Call, and Michael
  Tomasello.
\newblock A test of the submentalizing hypothesis: Apes' performance in a false
  belief task inanimate control.
\newblock {\em Communicative \& Integrative Biology}, 10(4):e1343771, 2017.

\bibitem{horschler2020non}
Daniel~J Horschler, Evan~L MacLean, and Laurie~R Santos.
\newblock Do non-human primates really represent others’ beliefs?
\newblock {\em Trends in Cognitive Sciences}, 24(8):594--605, 2020.

\bibitem{low2012implicit}
Jason Low and Josef Perner.
\newblock Implicit and explicit theory of mind: State of the art.
\newblock {\em British Journal of Developmental Psychology}, 30(1):1--13, 2012.

\bibitem{butterfill2013construct}
Stephen~A Butterfill and Ian~A Apperly.
\newblock How to construct a minimal theory of mind.
\newblock {\em Mind \& Language}, 28(5):606--637, 2013.

\bibitem{guss2019minerl}
William~H Guss, Brandon Houghton, Nicholay Topin, Phillip Wang, Cayden Codel,
  Manuela Veloso, and Ruslan Salakhutdinov.
\newblock Minerl: A large-scale dataset of minecraft demonstrations.
\newblock {\em arXiv preprint arXiv:1907.13440}, 2019.

\bibitem{wang2020enhanced}
Rui Wang, Joel Lehman, Aditya Rawal, Jiale Zhi, Yulun Li, Jeffrey Clune, and
  Kenneth Stanley.
\newblock Enhanced poet: Open-ended reinforcement learning through unbounded
  invention of learning challenges and their solutions.
\newblock In {\em International Conference on Machine Learning}, pages
  9940--9951. PMLR, 2020.

\bibitem{samvelyan2021minihack}
Mikayel Samvelyan, Robert Kirk, Vitaly Kurin, Jack Parker-Holder, Minqi Jiang,
  Eric Hambro, Fabio Petroni, Heinrich K{\"u}ttler, Edward Grefenstette, and
  Tim Rockt{\"a}schel.
\newblock Minihack the planet: A sandbox for open-ended reinforcement learning
  research.
\newblock {\em arXiv preprint arXiv:2109.13202}, 2021.

\bibitem{team2021open}
Open Ended~Learning Team, Adam Stooke, Anuj Mahajan, Catarina Barros, Charlie
  Deck, Jakob Bauer, Jakub Sygnowski, Maja Trebacz, Max Jaderberg, Michael
  Mathieu, et~al.
\newblock Open-ended learning leads to generally capable agents.
\newblock {\em arXiv preprint arXiv:2107.12808}, 2021.

\bibitem{hafner2021benchmarking}
Danijar Hafner.
\newblock Benchmarking the spectrum of agent capabilities.
\newblock {\em arXiv preprint arXiv:2109.06780}, 2021.

\bibitem{johnson2005subcortical}
Mark~H Johnson.
\newblock Subcortical face processing.
\newblock {\em Nature Reviews Neuroscience}, 6(10):766--774, 2005.

\bibitem{reid2017human}
V.~M. Reid, K.~Dunn, R.~J. Young, J.~Amu, T.~Donovan, and N.~Reissland.
\newblock {\em The human fetus preferentially engages with face-like visual
  stimuli}.
\newblock Current Biology, 2017.

\bibitem{farroni2005newborns}
Teresa Farroni, Mark~H Johnson, Enrica Menon, Luisa Zulian, Dino Faraguna, and
  Gergely Csibra.
\newblock Newborns' preference for face-relevant stimuli: Effects of contrast
  polarity.
\newblock {\em Proceedings of the National Academy of Sciences},
  102(47):17245--17250, 2005.

\bibitem{perszyk2018linking}
Danielle~R Perszyk and Sandra~R Waxman.
\newblock Linking language and cognition in infancy.
\newblock {\em Annual review of psychology}, 69:231--250, 2018.

\bibitem{ullman2012simple}
Shimon Ullman, Daniel Harari, and Nimrod Dorfman.
\newblock From simple innate biases to complex visual concepts.
\newblock {\em Proceedings of the National Academy of Sciences},
  109(44):18215--18220, 2012.

\bibitem{grossmann2017eyes}
Tobias Grossmann.
\newblock The eyes as windows into other minds: An integrative perspective.
\newblock {\em Perspectives on Psychological Science}, 12(1):107--121, 2017.

\bibitem{simion2008predisposition}
Francesca Simion, Lucia Regolin, and Hermann Bulf.
\newblock A predisposition for biological motion in the newborn baby.
\newblock {\em Proceedings of the National Academy of Sciences},
  105(2):809--813, 2008.

\bibitem{saxe2013theory}
Rebecca Saxe and Liane Young.
\newblock Theory of mind: How brains think about thoughts.
\newblock {\em The Oxford Handbook of Cognitive Neuroscience}, 2:204--213,
  2013.

\bibitem{koster2013decoding}
Jorie Koster-Hale, Rebecca Saxe, James Dungan, and Liane~L Young.
\newblock Decoding moral judgments from neural representations of intentions.
\newblock {\em Proceedings of the National Academy of Sciences},
  110(14):5648--5653, 2013.

\bibitem{wellman2004scaling}
Henry~M Wellman and David Liu.
\newblock Scaling of theory-of-mind tasks.
\newblock {\em Child Development}, 75(2):523--541, 2004.

\bibitem{matiisen2019teacher}
Tambet Matiisen, Avital Oliver, Taco Cohen, and John Schulman.
\newblock Teacher--student curriculum learning.
\newblock {\em IEEE Transactions on Neural Networks and Learning Systems},
  31(9):3732--3740, 2019.

\bibitem{forestier2017intrinsically}
S{\'e}bastien Forestier, R{\'e}my Portelas, Yoan Mollard, and Pierre-Yves
  Oudeyer.
\newblock Intrinsically motivated goal exploration processes with automatic
  curriculum learning.
\newblock {\em arXiv preprint arXiv:1708.02190}, 2017.

\bibitem{milligan2007language}
Karen Milligan, Janet~Wilde Astington, and Lisa~Ain Dack.
\newblock Language and theory of mind: Meta-analysis of the relation between
  language ability and false-belief understanding.
\newblock {\em Child Development}, 78(2):622--646, 2007.

\bibitem{hale2003influence}
Courtney~Melinda Hale and Helen Tager-Flusberg.
\newblock The influence of language on theory of mind: A training study.
\newblock {\em Developmental Science}, 6(3):346--359, 2003.

\bibitem{peterson2005steps}
Candida~C Peterson, Henry~M Wellman, and David Liu.
\newblock Steps in theory-of-mind development for children with deafness or
  autism.
\newblock {\em Child Development}, 76(2):502--517, 2005.

\bibitem{hofmann2016training}
Stefan~G Hofmann, Stacey~N Doan, Manuel Sprung, Anne Wilson, Chad Ebesutani,
  Leigh~A Andrews, Joshua Curtiss, and Paul~L Harris.
\newblock Training children’s theory-of-mind: A meta-analysis of controlled
  studies.
\newblock {\em Cognition}, 150:200--212, 2016.

\bibitem{devlin2018bert}
Jacob Devlin, Ming-Wei Chang, Kenton Lee, and Kristina Toutanova.
\newblock Bert: Pre-training of deep bidirectional transformers for language
  understanding.
\newblock {\em arXiv preprint arXiv:1810.04805}, 2018.

\bibitem{radford2019language}
Alec Radford, Jeffrey Wu, Rewon Child, David Luan, Dario Amodei, Ilya
  Sutskever, et~al.
\newblock Language models are unsupervised multitask learners.
\newblock {\em OpenAI Blog}, 1(8):9, 2019.

\bibitem{brown2020language}
Tom Brown, Benjamin Mann, Nick Ryder, Melanie Subbiah, Jared~D Kaplan, Prafulla
  Dhariwal, Arvind Neelakantan, Pranav Shyam, Girish Sastry, Amanda Askell,
  et~al.
\newblock Language models are few-shot learners.
\newblock {\em Advances in Neural Information Processing Systems},
  33:1877--1901, 2020.

\bibitem{fedus2021switch}
William Fedus, Barret Zoph, and Noam Shazeer.
\newblock Switch transformers: Scaling to trillion parameter models with simple
  and efficient sparsity, 2021.

\bibitem{baker2017rational}
Chris~L Baker, Julian Jara-Ettinger, Rebecca Saxe, and Joshua~B Tenenbaum.
\newblock Rational quantitative attribution of beliefs, desires and percepts in
  human mentalizing.
\newblock {\em Nature Human Behaviour}, 1(4):1--10, 2017.

\bibitem{jara2019theory}
Julian Jara-Ettinger.
\newblock Theory of mind as inverse reinforcement learning.
\newblock {\em Current Opinion in Behavioral Sciences}, 29:105--110, 2019.

\bibitem{gallese1998mirror}
Vittorio Gallese and Alvin Goldman.
\newblock Mirror neurons and the simulation theory of mind-reading.
\newblock {\em Trends in Cognitive Sciences}, 2(12):493--501, 1998.

\bibitem{rizzolatti2008mirrors}
Giacomo Rizzolatti and Corrado Sinigaglia.
\newblock {\em Mirrors in the brain: How our minds share actions and emotions}.
\newblock Oxford University Press, USA, 2008.

\bibitem{hickok2014myth}
Gregory Hickok.
\newblock {\em The myth of mirror neurons: The real neuroscience of
  communication and cognition}.
\newblock WW Norton \& Company, 2014.

\bibitem{heyes2010mirror}
Cecilia Heyes.
\newblock Where do mirror neurons come from?
\newblock {\em Neuroscience \& Biobehavioral Reviews}, 34(4):575--583, 2010.

\bibitem{heyes2022happened}
Cecilia Heyes and Caroline Catmur.
\newblock What happened to mirror neurons?
\newblock {\em Perspectives on Psychological Science}, 17(1):153--168, 2022.

\bibitem{Pomerleau1991}
Dean~A. Pomerleau.
\newblock Efficient training of artificial neural networks for autonomous
  navigation.
\newblock {\em Neural Computation}, 3(1):88--97, 1991.

\bibitem{Schaal1999}
Stefan Schaal.
\newblock Is imitation learning the route to humanoid robots?
\newblock {\em Trends in Cognitive Sciences}, 3(6):233--242, 1999.

\bibitem{Ng00algorithmsfor}
Andrew~Y. Ng and Stuart Russell.
\newblock Algorithms for inverse reinforcement learning.
\newblock In {\em in Proc. 17th International Conf. on Machine Learning}, pages
  663--670. Morgan Kaufmann, 2000.

\bibitem{christiano2017deep}
Paul Christiano, Jan Leike, Tom~B. Brown, Miljan Martic, Shane Legg, and Dario
  Amodei.
\newblock Deep reinforcement learning from human preferences, 2017.

\bibitem{wu2021recursively}
Jeff Wu, Long Ouyang, Daniel~M. Ziegler, Nisan Stiennon, Ryan Lowe, Jan Leike,
  and Paul Christiano.
\newblock Recursively summarizing books with human feedback, 2021.

\bibitem{lampinen2021tell}
Andrew~K. Lampinen, Nicholas~A. Roy, Ishita Dasgupta, Stephanie C.~Y. Chan,
  Allison~C. Tam, James~L. McClelland, Chen Yan, Adam Santoro, Neil~C.
  Rabinowitz, Jane~X. Wang, and Felix Hill.
\newblock Tell me why! -- explanations support learning of relational and
  causal structure, 2021.

\bibitem{abramson2021imitating}
Josh Abramson, Arun Ahuja, Iain Barr, Arthur Brussee, Federico Carnevale, Mary
  Cassin, Rachita Chhaparia, Stephen Clark, Bogdan Damoc, Andrew Dudzik, Petko
  Georgiev, Aurelia Guy, Tim Harley, Felix Hill, Alden Hung, Zachary Kenton,
  Jessica Landon, Timothy Lillicrap, Kory Mathewson, Soňa Mokrá, Alistair
  Muldal, Adam Santoro, Nikolay Savinov, Vikrant Varma, Greg Wayne, Duncan
  Williams, Nathaniel Wong, Chen Yan, and Rui Zhu.
\newblock Imitating interactive intelligence, 2021.

\bibitem{deepmindinteractiveagentsteam2022creating}
DeepMind Interactive~Agents Team, Josh Abramson, Arun Ahuja, Arthur Brussee,
  Federico Carnevale, Mary Cassin, Felix Fischer, Petko Georgiev, Alex Goldin,
  Mansi Gupta, Tim Harley, Felix Hill, Peter~C Humphreys, Alden Hung, Jessica
  Landon, Timothy Lillicrap, Hamza Merzic, Alistair Muldal, Adam Santoro, Guy
  Scully, Tamara von Glehn, Greg Wayne, Nathaniel Wong, Chen Yan, and Rui Zhu.
\newblock Creating multimodal interactive agents with imitation and
  self-supervised learning, 2022.

\bibitem{akula2022cx}
Arjun~R Akula, Keze Wang, Changsong Liu, Sari Saba-Sadiya, Hongjing Lu, Sinisa
  Todorovic, Joyce Chai, and Song-Chun Zhu.
\newblock Cx-tom: Counterfactual explanations with theory-of-mind for enhancing
  human trust in image recognition models.
\newblock {\em Iscience}, 25(1):103581, 2022.

\bibitem{edmonds2019tale}
Mark Edmonds, Feng Gao, Hangxin Liu, Xu~Xie, Siyuan Qi, Brandon Rothrock, Yixin
  Zhu, Ying~Nian Wu, Hongjing Lu, and Song-Chun Zhu.
\newblock A tale of two explanations: Enhancing human trust by explaining robot
  behavior.
\newblock {\em Science Robotics}, 4(37):eaay4663, 2019.

\bibitem{triesch2007emergence}
Jochen Triesch, Hector Jasso, and Gedeon~O De{\'a}k.
\newblock Emergence of mirror neurons in a model of gaze following.
\newblock {\em Adaptive Behavior}, 15(2):149--165, 2007.

\bibitem{du2019techniques}
Mengnan Du, Ninghao Liu, and Xia Hu.
\newblock Techniques for interpretable machine learning.
\newblock {\em Communications of the ACM}, 63(1):68--77, 2019.

\bibitem{olah2018building}
Chris Olah, Arvind Satyanarayan, Ian Johnson, Shan Carter, Ludwig Schubert,
  Katherine Ye, and Alexander Mordvintsev.
\newblock The building blocks of interpretability.
\newblock {\em Distill}, 3(3):e10, 2018.

\bibitem{molnar2020interpretable}
Christoph Molnar.
\newblock {\em Interpretable machine learning}.
\newblock Lulu. com, 2020.

\bibitem{alharbi2021learning}
Raed Alharbi, Minh~N Vu, and My~T Thai.
\newblock Learning interpretation with explainable knowledge distillation.
\newblock In {\em 2021 IEEE International Conference on Big Data (Big Data)},
  pages 705--714. IEEE, 2021.

\bibitem{hilton2020understanding}
Jacob Hilton, Nick Cammarata, Shan Carter, Gabriel Goh, and Chris Olah.
\newblock Understanding rl vision.
\newblock {\em Distill}, 5(11):e29, 2020.

\bibitem{alain2016understanding}
Guillaume Alain and Yoshua Bengio.
\newblock Understanding intermediate layers using linear classifier probes.
\newblock {\em arXiv preprint arXiv:1610.01644}, 2016.

\bibitem{meyes2019ablation}
Richard Meyes, Melanie Lu, Constantin~Waubert de~Puiseau, and Tobias Meisen.
\newblock Ablation studies in artificial neural networks.
\newblock {\em arXiv preprint arXiv:1901.08644}, 2019.

\bibitem{jaderberg2019human}
Max Jaderberg, Wojciech~M Czarnecki, Iain Dunning, Luke Marris, Guy Lever,
  Antonio~Garcia Castaneda, Charles Beattie, Neil~C Rabinowitz, Ari~S Morcos,
  Avraham Ruderman, et~al.
\newblock Human-level performance in 3d multiplayer games with population-based
  reinforcement learning.
\newblock {\em Science}, 364(6443):859--865, 2019.

\bibitem{cichy2019deep}
Radoslaw~M Cichy and Daniel Kaiser.
\newblock Deep neural networks as scientific models.
\newblock {\em Trends in cognitive sciences}, 23(4):305--317, 2019.

\end{thebibliography}
\end{document}